%% file: CVPR2023_ASTR.tex
\documentclass[10pt,twocolumn,letterpaper]{article}

\usepackage[pagenumbers]{cvpr} 

\usepackage{graphicx}
\usepackage{amsmath}
\usepackage{amssymb}
\usepackage{booktabs}

\usepackage{amsthm}
\usepackage{multirow}

\usepackage[pagebackref,breaklinks,colorlinks]{hyperref}
\usepackage[accsupp]{axessibility}

\usepackage[capitalize]{cleveref}
\crefname{section}{Sec.}{Secs.}
\Crefname{section}{Section}{Sections}
\Crefname{table}{Table}{Tables}
\crefname{table}{Tab.}{Tabs.}

\def\cvprPaperID{19} 
\def\confName{CVPR}
\def\confYear{2023}

\begin{document}

\title{Structured Epipolar Matcher for Local Feature Matching}


\author{Jiahao Chang\footnotemark[1], Jiahuan Yu\footnotemark[1], Tianzhu Zhang\footnotemark[2]\\
{\small University of Science and Technology of China}
}

\maketitle

\renewcommand{\thefootnote}{\fnsymbol{footnote}} 
\footnotetext[1]{Equal contribution} 
\footnotetext[2]{Corresponding author} 

\begin{abstract}
\input{abstract}
\end{abstract}

\section{Introduction}
\label{sec:intro}
\input{intro}

\section{Related Work}
\label{sec:rw}
\input{rw}

\section{Our Approach}
\label{sec:method}
\input{method}
\section{Experiments}
\label{sec:expr}
\input{expr}

\section{Conclusion}
\label{sec:conclusion}
\input{conclusion}

{\small
\bibliographystyle{ieee_fullname}
\bibliography{CVPR2023bib}
}

\end{document}

%% file: abstract.tex
Local feature matching is challenging due to textureless and repetitive patterns.
Existing methods focus on using appearance features and global interaction and matching, while the importance of geometry priors in local feature matching has not been fully exploited.
Different from these methods, in this paper, we delve into the importance of geometry prior and propose Structured Epipolar Matcher (SEM) for local feature matching, which can leverage the geometric information in an iterative matching way.
The proposed model enjoys several merits.
First, our proposed Structured Feature Extractor can model the relative positional relationship between pixels and high-confidence anchor points.
Second, our proposed Epipolar Attention and Matching can filter out irrelevant areas by utilizing the epipolar constraint.
Extensive experimental results on five standard benchmarks demonstrate the superior performance of our SEM compared to state-of-the-art methods.
Project page: \url{https://sem2023.github.io}.

%% file: intro.tex
Local feature matching, which aims to find correspondences between a pair of images, is essential for many important tasks in computer vision, such as Structure-from-Motion (SfM)~\cite{schonberger2016structure}, 3D reconstruction~\cite{dai2017bundlefusion}, visual localization~\cite{sattler2018benchmarking, taira2018inloc}, and pose estimation~\cite{grabner20183d, persson2018lambda}.
Given its broad application, local feature matching has received significant attention, leading to the development of many research studies.
Despite this, accurate local feature matching remains a challenging task, particularly in scenarios with poor texture, repetitive texture, illumination variations, and scale changes.

Numerous methods~\cite{detone2018superpoint, li20dualrc, r2d2, rocco2018neighbourhood, sun2021loftr} have been proposed to overcome the challenges mentioned above, which can be divided into two major groups: detector-based methods~\cite{barroso2019key, detone2018superpoint, dusmanu2019d2, ono2018lf, r2d2, sarlin2020superglue} and detector-free methods~\cite{huang2019dynamic, li20dualrc, rocco2020efficient, rocco2018neighbourhood, sun2021loftr, chen2022aspanformer,yu2023adaptive,edstedt2022deep}.
The detector-based approach consists of three separate stages: first, extracting keypoints from the images; then, generating descriptors of the keypoints; and finally, finding correspondences between keypoints using a matcher.
The performance of the detector-based method highly relies on the quality of keypoints.
Unfortunately, reliable keypoint detection in textureless or repetitive texture areas is a highly challenging problem that limits the final performance of detector-based methods.
In comparison, detector-free approaches build dense correspondences between pixels instead of extracted keypoints.
Recent works show that this method can handle poor textured regions better and demonstrate excellent performance.
Some recent works~\cite{sun2021loftr,jiang2021cotr,chen2022aspanformer,wang2022matchformer} leverage the Transformer architecture and shows impressive performance.
However, most of these works use an all-pixel-to-all-pixel matching process, which introduces irrelevant pixels and negatively affects the result in some scenes, such as regions with repetitive textures.

\begin{figure}[t]
    \centering
    \includegraphics[width=1.0\linewidth]{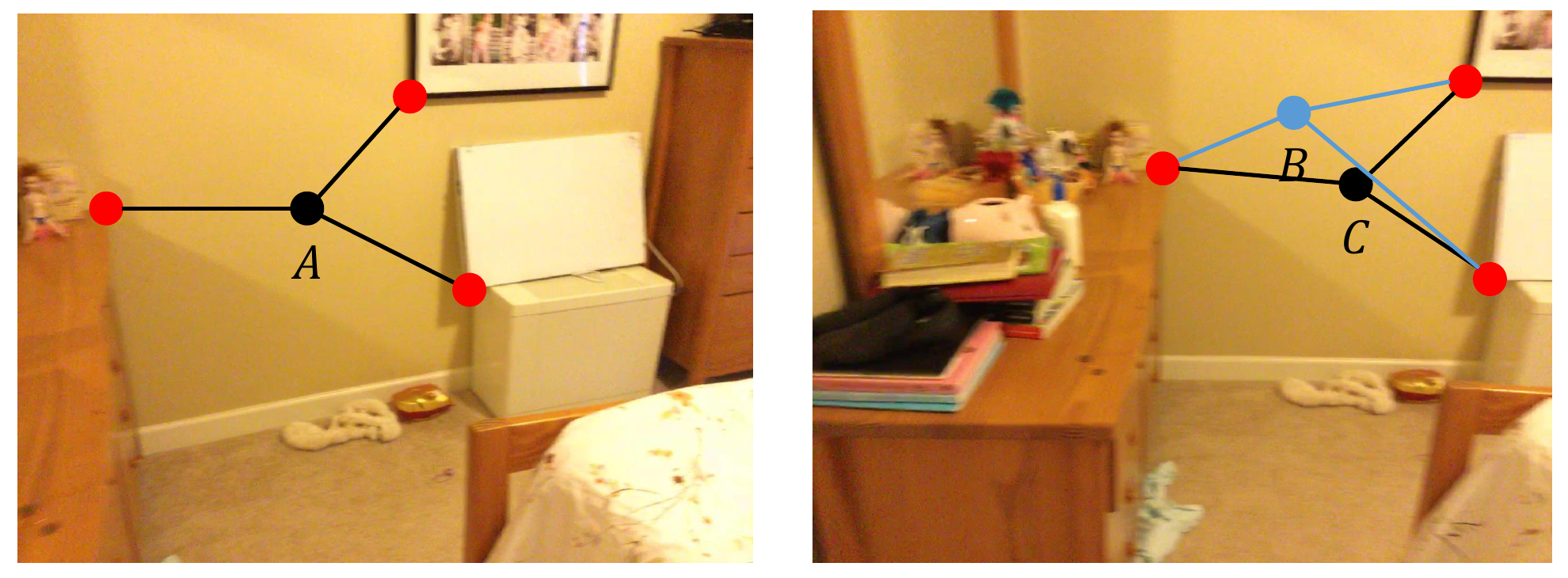}
    \caption{
    With the help of anchor points (marked in {\color{red} red}), we can correctly match point $A$ to point $C$ and remove interference from point $B$.
    }\label{fig:motivation_s}
    \vspace{-3 mm}
\end{figure}

\begin{figure}[t]
    \centering
    \includegraphics[width=1.0\linewidth]{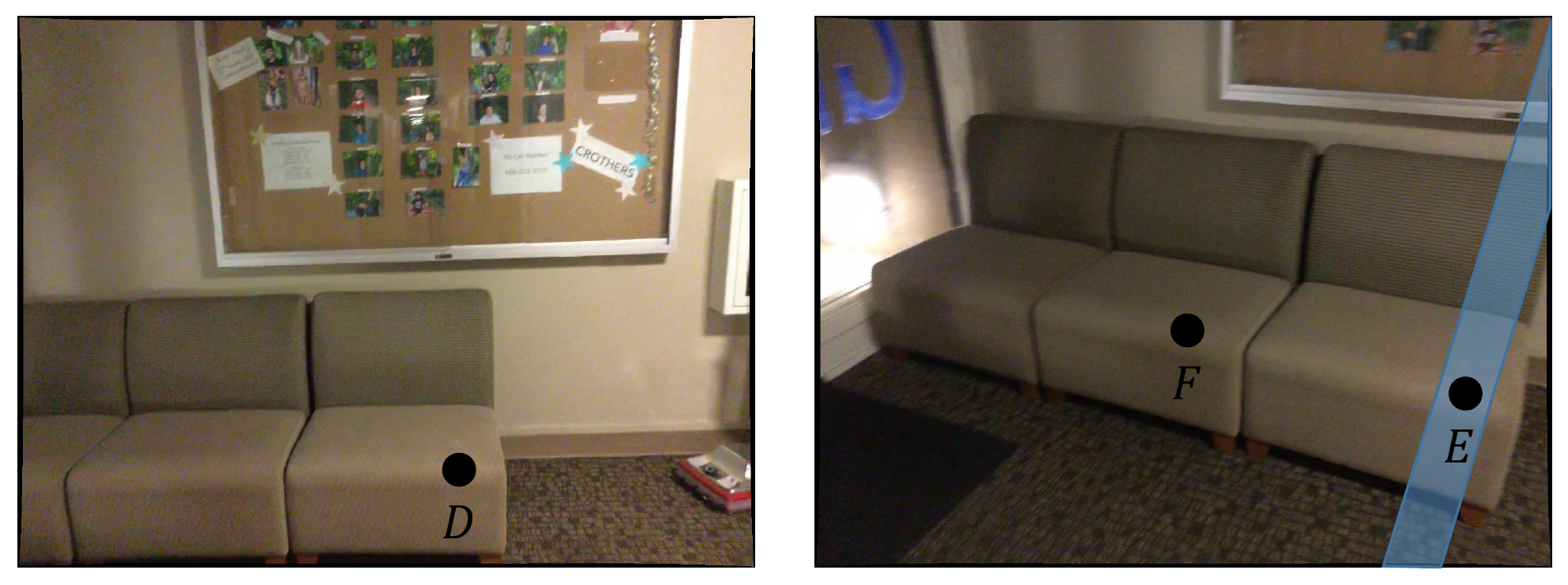}
    \caption{
    Using the epipolar geometry prior to calculate the candidate matching area of point $D$, we can exclude the influence of point $F$ on $D$.
    }\label{fig:motivation_e}
    \vspace{-3 mm}
\end{figure}

Upon studying the previous matching methods, we have identified two issues that cannot be ignored in obtaining dense correspondences between images:
(1) \textbf{How to extract more distinguishable features in areas with poor or repetitive textures.}
Existing methods focus on extracting features from appearance, which is not distinguishable enough.
As shown in motivation Figure~\ref{fig:motivation_s}, points $A$, $B$ and $C$ are similar in appearance, making it difficult to obtain accurate correspondence.
In contrast, textured anchor points (marked in {\color{red} red}) can  easily match with the correct correspondence through appearance.
We have discovered an interesting fact that by using the relative positional relationship between point $A$ and other anchor points, we can easily determine the correct matching position of $A$.
Therefore, it is necessary to consider the relative positional relationship with the anchor point during matching, which can help us make better judgments.
(2) \textbf{How to filter out irrelevant regions as much as possible during the matching process.}
Existing methods usually use Transformers for global feature update and matching, which would make matching process influenced by unrelated areas.
As shown in motivation Figure~\ref{fig:motivation_e}, the correct match of point $D$ is point $E$.
If point $D$ aggregates the features of point $F$, it will confuse the network to determine the final matching result.
Therefore, it is necessary to design a way to filter out the negative influence of irrelevant areas such as $F$.

To address the aforementioned issues, we propose the \textbf{Structured Epipolar Matcher} (SEM) with an \textbf{Iterative Epipolar Coarse Matching} stage, which consists of a \textbf{Structured Feature Extractor} and \textbf{Epipolar Attention/Matching Module}.
In the Structured Feature Extractor, we select high-confidence matching pixels as anchor points, and encode the relative position between each point of the image and the anchor points in a scale and rotation invariant manner as a supplement to the appearance features, making the features more discriminating.
In the Epipolar Attention/Matching Module, we consider the epipolar geometry prior to filter out candidate matching regions.
First we select points with high confidence, which is used to calculate the relative pose of the camera. 
And then, we use the epipolar area corresponding to each point as the valid area (painted in {\color{blue} blue} in Figure~\ref{fig:motivation_e}) for feature interaction and matching.
The entire Iterative Epipolar Coarse Matching process is iterated to gradually optimize the matching results.

The contributions of our method could be summarized into three-fold:
(1) We proposed a novel Structured Epipolar Matcher (SEM), which includes a Structured Feature Extractor and Epipolar Attention/Matching Module in a unified architecture.
(2) We design the Structured Feature Extractor, which can generate structured feature to complement the appearance features, making them more discriminating.
Also, we design Epipolar Attention/Matching Module, which uses epipolar geometry prior to filter out irrelevant matching regions as much as possible, and Iterative Epipolar Coarse Matching process will gradually optimize the matching results.
(3) Extensive experimental results on four challenging benchmarks show that our proposed method performs favorably against state-of-the-art image matching methods.

%% file: rw.tex
This section provides a brief overview of the methods related to local feature matching.

\noindent\textbf{Local Feature Matching}.
The popular local feature matching methods can be divided into two paradigms: detector-based methods and detector-free methods.
Detector-based methods involve three main stages: feature detection, feature description, and feature matching.
Hand-crafted local features such as SIFT~\cite{lowe2004distinctive} and ORB~\cite{rublee2011orb} are the most widely used.
Recently, learning-based local features like LIFT~\cite{yi2016lift} and SuperPoint~\cite{detone2018superpoint} have been shown to significantly improve performance compared to classical methods.
Some researchers also focus on enhancing the feature matching stage, representative works include D2Net~\cite{dusmanu2019d2}, R2D2~\cite{r2d2} and SuperGlue~\cite{sarlin2020superglue}.
Nevertheless, detector-based methods rely on keypoint detectors, which limits the performance in challenging scenarios such as repetitive textures, weak textures, and illumination changes.
On the contrary,  Detector-free approaches directly match dense features between pixels without the local feature detector.
While classical methods~\cite{lucas1981iterative,horn1981determining} exist, few of them outperform detector-based methods.
Learning-based methods have revolutionized this field, with cost-volume-based methods~\cite{rocco2018neighbourhood,li20dualrc,truong2020glu,truong2020gocor,truong2021pdc} and Transformer-based methods~\cite{jiang2021cotr,sun2021loftr,wang2022matchformer,chen2022aspanformer,yu2023adaptive} leading the charge.
While cost-volume-based methods archive promising performance, the limited perceptual field of CNNs is its inescapable shortcoming.
Recently, Transformer-based methods overcome this problem and leads the benchmark.
Given that detector-free methods have been shown to perform better in local feature matching, we have chosen this paradigm as our baseline approach, and use our proposed new module to overcome the challenging scenarios.

\noindent\textbf{Geometry Prior in Local Feature Matching.}
The significance of geometric priors in local feature matching has been explored in previous works.
GeoWrap~\cite{berton2021viewpoint} uses a learned pairwise warping to increase visual overlap between images.
Toft et al. \cite{toft2020single} propose a method to remove perspective distortion based on single-image depth estimation.
RoRD~\cite{parihar2021rord} generate rotation-robust descriptors via orthographic viewpoint projection.
COTR~\cite{jiang2021cotr} also estimates scale by finding co-visible region, and uses recursive zooming to enable the matcher to sense geometry scale.
Patch2Pix~\cite{zhou2021patch2pix} uses the consistent with the epipolar geometry of an input image pair to supervise the training.
In our research, we delve deeper into the role of geometric prior in local feature matching and propose a novel and effective method to leverage epipolar constraint and relative position information in handling challenging scenes.

\noindent\textbf{Iterative Optimization.}
Iterative optimization is a common paradigm in computer vision and has been explored in many previous work~\cite{sun2020acne,sun2023neuralbf,teed2020raft,jiang2021cotr,truong2020glu,truong2021pdc}.
For example, RAFT~\cite{teed2020raft} iteratively updates a flow field through a recurrent unit that performs lookups on the correlation volumes, ACNe~\cite{sun2020acne} uses an robust iterative optimization to build permutation-equivariant network, and NeuralBF~\cite{sun2023neuralbf} proposes an iterative bilateral filtering with learned kernels for instance segmentation on point clouds.
The iterative optimization is also deployed by some works in local feature matching, such as PDC-Net~\cite{truong2021pdc}, GLU-Net~\cite{truong2020glu} and COTR~\cite{jiang2021cotr}.
In this work, we follow this tried and tested paradigm and combine it with geometry priors.

%% file: method.tex
\begin{figure*}[t]
    \centering
    \includegraphics[width=1.0\linewidth]{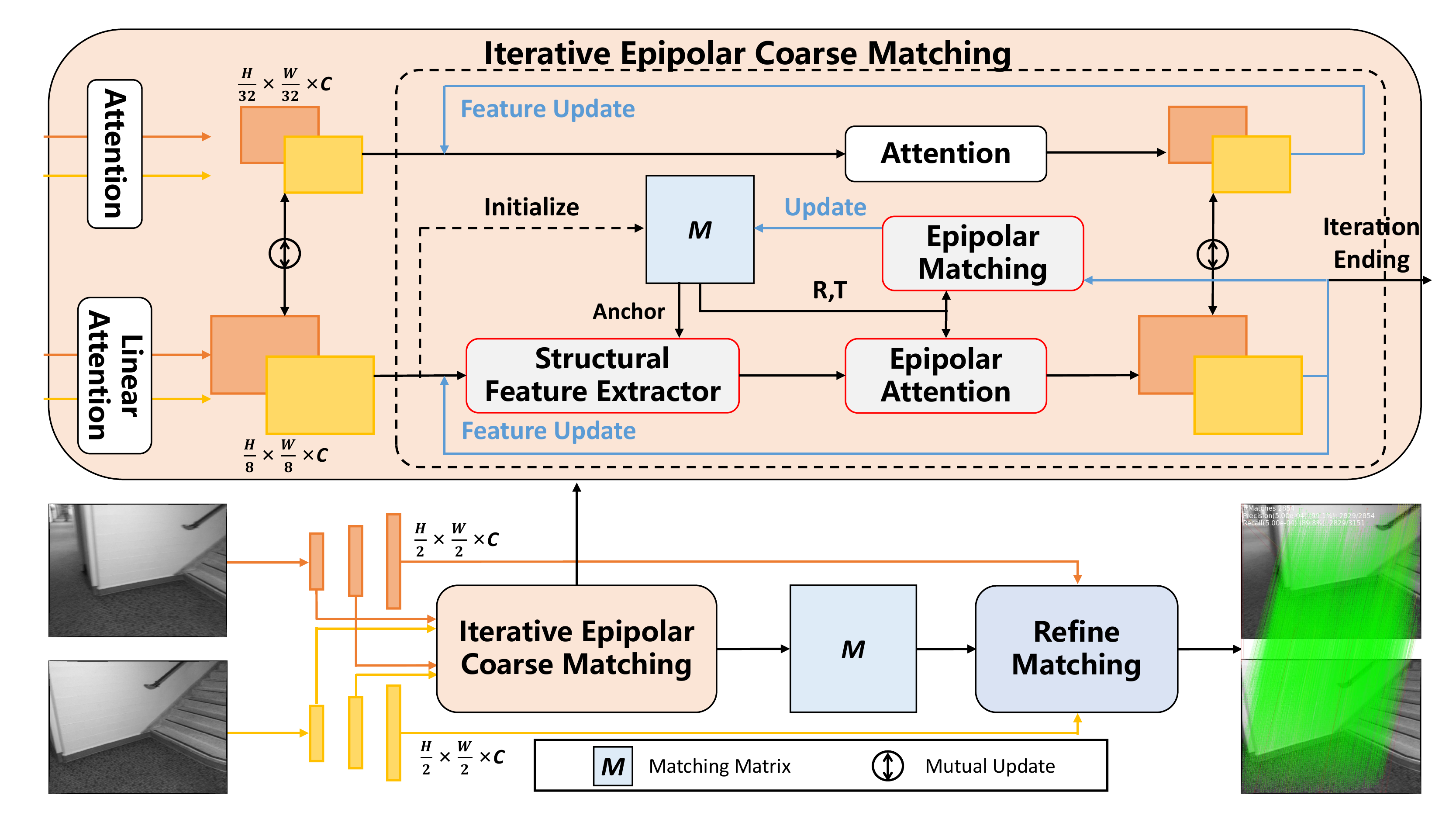}
    \caption{
    The architecture of our SEM.
    Our SEM consists of two stages: Iterative Epipolar Coarse Matching stage and refine Matching stage.
    Here, ``Attention'' means vanilla self-and-cross attention, ``Linear Attention'' means linear self-and-cross attention.
    Please refer to the text for detailed architecture.
    }\label{fig:method}
    \vspace{-3 mm}
\end{figure*}
In this section, we introduce our proposed Structured Epipolar Matcher (SEM) for local feature matching.
The overall architecture is illustrated in Figure~\ref{fig:method}.
Here we first present the overall scheme of SEM in Section~\ref{sec:3.1}.
In Section~\ref{sec:3.2}, we describe our Iterative Epipolar Coarse Matching in detail.
Then we explain our Structured Feature Extractor in Section~\ref{sec:3.3} and our Epipolar Attention/Matching in Section~\ref{sec:3.4}.
At last, Section~\ref{sec:3.5} states the loss used in our SEM. 

\subsection{Overview}\label{sec:3.1}
As shown in Figure~\ref{fig:method}, the proposed SEM mainly consists of two stages, including a Iterative Epipolar Coarse Matching stage and refine Matching stage.
Here we provide a concise overview of the complete process.
For a pair of input images, reference image $I_{ref}$ and source image $I_{src}$, we first use a Feature Pyramid Network (FPN)~\cite{lin2017feature} to extract multi-level features of each image $\{f_{1/i}^{ref},f_{1/i}^{src}\}$.
Then the feature maps with the size of ${1/32}$ and ${1/8}$ are sent into the Iterative Epipolar Coarse Matching stage.
With the help of ${1/32}$ feature maps, coarse matching matrix $M$ are obtained from the feature maps with the size of ${1/8}$ through a Structured Feature Extractor, Epipolar Attention and Epipolar Matching.
Finally, we refine our coarse matching results in refine matching, which is the same as LoFTR~\cite{sun2021loftr}.

\subsection{Iterative Epipolar Coarse Matching}\label{sec:3.2}
Considering that global feature can help fine feature to locate correct matching area, we use a multi-level strategy~\cite{yu2023adaptive} in Iterative Epipolar Coarse Matching.
First, we apply one self-and-cross attention layer on the feature maps $\{f_{1/32}^{ref},f_{1/32}^{src}\}$ and one self-and-cross linear attention layer on the feature maps $\{f_{1/8}^{ref},f_{1/8}^{src}\}$.
After that, features at two scales are updated from each other. 
This progress can be described as following:
\begin{equation}\label{eq:epc0}
    \begin{aligned}
    &\hat{f}_{1/32}^{t} = f_{1/32}^{t} + \mathrm{Conv_{1 \times 1}}(\mathrm{Down}(f_{1/8}^{t})), t \in \{ref, src\}
    \\
    &\hat{f}_{1/8}^{t} = f_{1/8}^{t} + \mathrm{Conv_{1 \times 1}}(\mathrm{Up}(f_{1/32}^{t})), t \in \{ref, src\}
    \end{aligned}
\end{equation}
where $\hat{f}_{1/32}^{t}$ and $\hat{f}_{1/8}^{t}$ denotes the fused features with the size of ${1/32}$ and ${1/8}$.
The purpose of the initialization is to establish the global connection in image or cross images.
Next, we enter the process of iterative matching.
During iterative process, $f_{1/32}^{ref}$ and $f_{1/32}^{src}$ are updated by self-and-cross attention layers to mainatin the global connection.
Coarse matching matrix $M \in \mathbb{R}^{HW \times HW}$ is initialized by computing pointwise similarity from flattened $f_{1/8}^{ref}$ and $f_{1/8}^{src}$ and a dural-softmax operator~\cite{sun2021loftr}:
\begin{equation}\label{eq:epc1}
    \begin{aligned}
    &S(i,j) = \tau \langle f_{1/8}^{ref}(i), f_{1/8}^{src}(j) \rangle,
    \\
    &\noindent M(i, j) = \mathrm{softmax}(S(i, :))(i,j) \cdot \mathrm{softmax}(S(:, j))(i,j),
    \end{aligned}
\end{equation}
 where $i$ is a pixel in image $I_{ref}$ and $j$ is a pixel in image $I_{src}$.
 $\tau$ means the temperature coefficient.
 $S \in R^{HW \times HW}$ is the similarity matrix.
As features on appearance, $f_{1/8}^{ref}$ and $f_{1/8}^{src}$ are supplied by structured feature from Structured Feature Extractor and become more discriminating.
And then, in Epipolar Attention, we interact $f_{1/8}^{ref}$ and $f_{1/8}^{src}$ via cross attention within corresponding epipolar banded area, which will be detailed in Section~\ref{sec:3.4}.
After updating features between different scales, in Epipolar Matching, we use $f_{1/8}^{ref}$ and $f_{1/8}^{src}$ to rewrite matching matrix $M$ along epipolar banded areas, which is a sparse similarity calculation with a low computational cost.
After several iterations of epipolar matching, we obtain accurate coarse matching results.

\begin{figure}[t]
    \centering
    \includegraphics[width=1.0\linewidth]{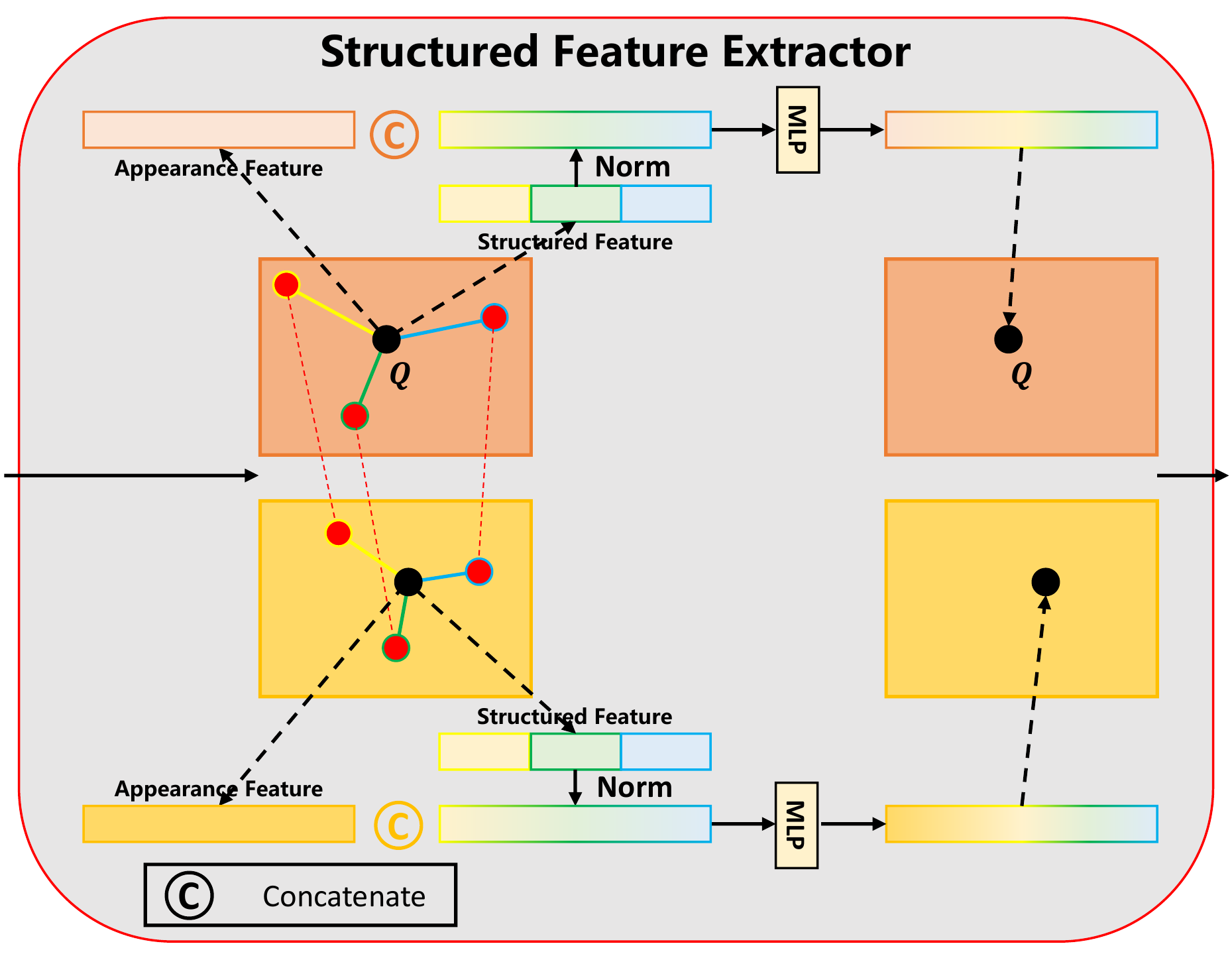}
    \caption{
    The schematic of our Structural Feature Extractor.
    We take the black point to examplify the construction of our structural feature, and we show the way to fuse appearance feature and structural feature.
    The {\color{red} red} points means the selected anchor points.
    }\label{fig:Structural Feature Extractor}
    \vspace{-3 mm}
\end{figure}

\subsection{Structured Feature Extractor}\label{sec:3.3}
As human beings, when we search for correspondences between a pair of images, we do not just compare pixels.
Instead, we first look for conspicuous and easily matched anchor points in the images.
By utilizing the relative positional relationships between pixels and anchor points, human beings can easily overcome the challenges posed by textureless and repetitive texture areas.
This process is also inspiring for learning-based approaches.
As shown in Figure~\ref{fig:motivation_s}, even though the appearance features of points $B$ and $C$ are not discriminating enough, suitable anchor points can be selected, and the relative positions can be utilized to build structured features, which can compensate for the shortage of appearance features, resulting in better correspondences in textureless and repetitive texture regions.

We propose the \textbf{Structured Feature Extractor} for above purpose.
First we define the confidence of correspondence:
\begin{equation}\label{eq:def_of_conf}
\mathrm{conf}(p_i)=\max \left\{ M (i, j) \mid j\in\{1,2,\cdots,HW\} \right\},
\end{equation}
where $p_i$ is the pixel in reference image.
As shown in Figure~\ref{fig:Structural Feature Extractor}, given two image feature maps, $f^{ref}$ and $f^{src}$, we first refer to matching matrix $M$ to find high-confidence correspondence over a threshold of $\sigma_h$.
Among these correspondences, we randomly select $N_A$ pairs as anchor points (marked in red), denote as 
\begin{equation}\label{eq:def_of_anchor}
\begin{aligned}
 & \mathcal{A}^{ref}=\left\{ \left( x^{ref}_i, y^{ref}_i \right) \mid i=1,2,\cdots,N_A \right\}, \\
 & \mathcal{A}^{src}=\left\{ \left( x^{src}_i, y^{src}_i \right) \mid i=1,2,\cdots,N_A \right\}. \\
\end{aligned}
\end{equation}
Without loss of generality, we talk about the reference feature map $f^{ref}$. For any point $Q$ in reference image whose coordinate is $(x,y)$, we calculate the coordinate differences and Euclidean distances from $Q$ to all anchor points:
\begin{equation}\label{eq:p_anchor_diff}
\begin{aligned}
\Delta X^{ref} &= \left(x-x^{ref}_1,x-x^{ref}_2,\cdots,x-x^{ref}_{N_A} \right), \\
\Delta Y^{ref} &= \left(y-y^{ref}_1,y-y^{ref}_2,\cdots,y-y^{ref}_{N_A} \right), \\
D^{ref} &= \sqrt{\left(\Delta X\right)^2+ \left(\Delta Y\right)^2} = \left( d^{ref}_1, d^{ref}_2,\cdots,d^{ref}_{N_A} \right). \\
\end{aligned}
\end{equation}
The structured feature $f^{ref}_{sf}$ is then defined as
\begin{equation}\label{eq:def_sf}
\begin{aligned}
    & \Delta X^{ref}_n = \mathrm{norm} \left(\Delta X^{ref}\right), \\
    & \Delta Y^{ref}_n = \mathrm{norm} \left(\Delta Y^{ref}\right), \\
    & D^{ref}_n = \mathrm{norm} \left(\Delta D^{ref}\right), \\
    & f^{ref}_{sf}(x,y) = \Delta X^{ref}_n \Vert \Delta Y^{ref}_n \Vert D^{ref}_n, \\
\end{aligned}
\end{equation}
where $\Vert$ is concatenate operation, $\mathrm{norm}$ is the $L_1$ normalization operation. $f^{src}_{sf}(x,y)$ is calculated by similar process.

\begin{figure}[t]
    \centering
    \includegraphics[width=1.0\linewidth]{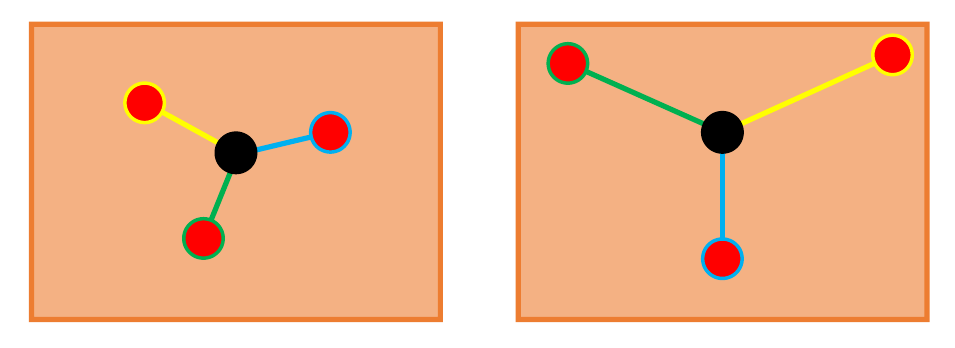}
    \caption{
    After scaling and rotating the image, the relative positional relationship between the anchor points (marked as {\color{red} red}) and the given point (marked as black).
    }\label{fig:norm}
    \vspace{-3 mm}
\end{figure}

Why do we need $D$ and $L_1$ normalization to model the relative position instead of only using $\Delta X$ and $\Delta Y$?
A suitable structured feature should satisfy both \textbf{scaling invariance} and \textbf{rotational invariance}.
As shown in Figure~\ref{fig:norm}, when an image is resized, $\Delta X$, $\Delta Y$ and $D$ are all scaled proportionally.
A $L_1$ normalization can solve the problems caused by scaling variation and make the structured feature more robust.
Also, considering the rotation of the images, $\Delta X$ and $\Delta Y$ will also change despite the normalization.
This is the motivation why we introduce $D$, which is rotational invariant with the normalization. 

Finally, the fused feature $f^{src}_{fused}$ can be obtained by applying a MLP to fuse the appearance feature and structural feature:
\begin{equation}\label{eq:feat_fusing}
f^{src}_{fused}(x,y)=\mathrm{MLP}\left( f^{src}(x,y) \Vert f^{src}_{sf}(x,y) \right)
\end{equation}

\begin{figure}[t]
    \centering
    \includegraphics[width=1.0\linewidth]{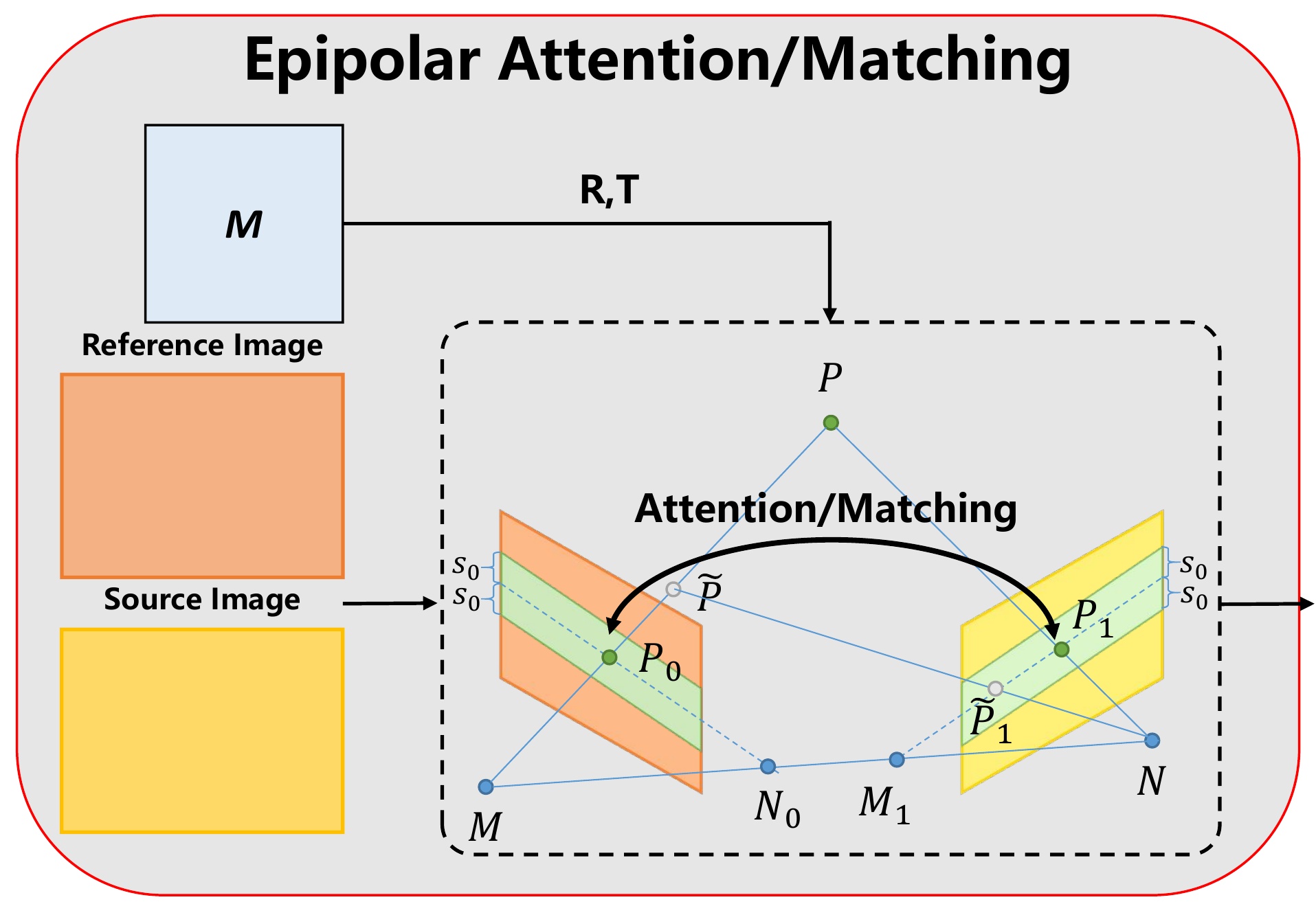}
    \caption{
    The illustration of our Epipolar Attention and Epipolar Matching.
    }\label{fig:Epipolar_Matching}
    \vspace{-3 mm}
\end{figure}

\subsection{Epipolar Attention and Matching}\label{sec:3.4}
Existing methods mostly perform global interaction between reference and source images, which will introduce irrelevant noisy points.
In Figure~\ref{fig:motivation_e}, the point $F$ has an negative impact on the matching result of point $D$.
Therefore, we attempt to remove irrelevant regions during feature updating and matching as soon as possible.
But how?
We think of some ``good" points that are easily to match correctly.
With the help of epipolar geometry prior, these ``good" points can filter out irrelevant areas and reduce candidate regions for other points.
Next, we will introduce the preliminaries of attention mechanism and epipolar line in Section~\ref{sec:3.4.1}.
Finally, we states our epipolar attention and matching in Section~\ref{sec:3.4.2}.

\subsubsection{Attention and Epipolar Line}\label{sec:3.4.1}

Vanilla attention calculation can be devided into 3 steps.
Inputs are converted to query $Q$, key $K$ and value $V$ by MLPs.
To start with, the attention map is computed by dot production from query $Q$ and key $K$.
After the softmax operator, the attention map becomes the weight matrix. 
Finally, the output is weighted sum of value $V$ with the weight matrix above.
The whole process can be expressed as 
\begin{equation}\label{eq:epc3}
    \mathrm{Attention}(Q, K, V) = \mathrm{softmax}(QK^T)V.
\end{equation}
The attention map can produce high computational and memory consumption, which is overwhelming for computer vision tasks.
To resolve the problem above, Katharopoulos~\cite{katharopoulos2020transformers} propose linear attention, which approximates the softmax operator with the product of two kernel functions $\phi(\cdot) = \mathrm{elu}(\cdot) + 1$ and uses the commutative law of matrix multiplication:
\begin{equation}\label{eq:epc4}
    \mathrm{Linear\_attention}(Q, K, V) = \phi (Q) (\phi (K^T)V),
\end{equation}
However, some experiments~\cite{germain2021visual,chen2021learning} have proved that linear attention will cause a slight decline in model performance.

In Figure~\ref{fig:Epipolar_Matching}, $P$ is a point in the real world, $P_0$ and $P_1$ is a pair of matching points, which are also the projections of $P$ on the reference and source image.
$M$ and $N$ are optical centers of the reference and source camera.
$N_0$ is the projection of $N$ on the reference image.
$M_1$ is the projection of $M$ on the source image.
$N_0$ and $M_1$ are a pair of epipolar points.
$P_0N_0$ and $P_1M_1$ are a pair of epipolar lines.
According to epipolar constraint, for any point in $P_0N_0$, its matching point on the source image will definitely fall on $P_1M_1$.

\subsubsection{Epipolar Attention and Matching}\label{sec:3.4.2}
In Figure~\ref{fig:Epipolar_Matching}, given Mating Matrix $M$, we first pick up ``good" matching pairs whose confidence is above a threshold $\sigma_h$.
With these matching pairs, we can obtain the relative pose of two images $\{R,T\}$.
Without loss of generality, we take the point $P_0$ as an example to show the way to find the candidate matching area of $P_0$.
The coordinate of $P_0$ on the reference image is $(x,y)^T$.
Then we find a point $\widetilde{P}$ on $MP$ whose reference camera coordinate is $K_{ref}^{-1} \cdot (x,y,1)^T$, where $K_{ref}$ denotes the intrinsics of the reference camera.
And the source camera coordinate of $\widetilde{P}$ is $R \cdot K_{ref}^{-1} \cdot (x,y,1)^T + T$.
Project $\widetilde{P}$ to the source image plane and we get $\widetilde{P}_1$.
The coordinate of $\widetilde{P}_1$ on the source image $c_{\widetilde{P}_1}$is 
\begin{equation}\label{eq:epc5}
\begin{aligned}
& p_{\widetilde{P}_1} = K_{src} \cdot (R \cdot K_{ref}^{-1} \cdot (x,y,1)^T + T),
\\
& c_{\widetilde{P}_1} = (\frac{p_{\widetilde{P}_1}(1)}{p_{\widetilde{P}_1}(2)}, \frac{p_{\widetilde{P}_1}(0)}{p_{\widetilde{P}_1}(2)}),
\end{aligned}
\end{equation}
where $K_{src}$ is the intrinsics of the source camera.
Meanwhile, the coordinate of $M_1$ on the source image $c_{M_1}$is 
\begin{equation}\label{eq:epc6}
\begin{aligned}
& p_{M_1} = K_{src} \cdot T,
\\
& c_{M_1} = (\frac{p_{M_1}(1)}{p_{M_1}(2)}, \frac{p_{M_1}(0)}{p_{M_1}(2)}).
\end{aligned}
\end{equation}
Now we can easily get the slope $k$ and intercept $b$ of $M_1\widetilde{P}_1$:
\begin{equation}\label{eq:epc7}
\begin{aligned}
& k = \frac{p_{\widetilde{P}_1}(1) - p_{M_1}(1)}{p_{\widetilde{P}_1}(0) - p_{M_1}(0)},
\\
& b = p_{M_1}(1) - k \cdot p_{M_1}(0).
\end{aligned}
\end{equation}
Considering that the initial relative pose $\{R,T\}$ may be inaccurate, we expand the epipolar region to a banded region with tolerance $s_0$, whose boundaries can be represented as $\{k,b - s_0\}$ and $\{k,b + s_0\}$.
At this point, we finally locate the candidate matching area of $P_0$.
The candidate matching area of points on the source image can be obtained in the similar way.
Each point performs cross attention with points in its corresponding candidate region.
For later updating matching matrix $M$, each point is also matched against points in its corresponding candidate region.

\subsection{Loss Function}\label{sec:3.5}

Our loss function mainly includes two parts, iterative coarse matching loss and refine matching loss.
Iterative coarse matching loss $L_i$ is mainly used to supervise the matching matrix $M$ updated by each iteration:
\begin{equation}\label{eq:epc11}
\noindent L_i = \sum_{k} - \frac{1}{\left\lvert M^{gt}_c \right\rvert} \sum_{(i,j) \in M^{gt}_c} \log M^k(i,j),
\end{equation}
where $M^{gt}_c$ is the ground truth matches at coarse resolution.
$M^k(i,j)$ denotes the predicted matching matrix in the $k$-th iteration.
Fine matching loss $L_f$ is a weighted $L_2$ loss same as LoFTR~\cite{sun2021loftr}:
\begin{equation}\label{eq:epc13}
\noindent L_f = \frac{1}{\left\lvert M^{gt}_f \right\rvert} \sum_{(i,j) \in M^{gt}_f} \frac{1}{\sigma ^2(i)}{\left\lVert j - j_{gt}\right\rVert}_2 ,
\end{equation}
where $M^{gt}_f$ is the ground truth matches at fine resolution and $\sigma^2(i)$ means the total variance of the heatmap.
Our final loss is:
\begin{equation}\label{eq:epc14}
\noindent L_{total} = L_i + L_f.
\end{equation}

%% file: expr.tex
This section presents the comprehensive evaluation of our SEM through a series of experiments.
To begin with, we describe the implementation details, and then we conduct experiments on four widely-used standard benchmarks for image matching.
Moreover, a range of ablation studies are conducted to validate the effectiveness of each individual component.

\subsection{Implementation Details}\label{4.1}

Our SEM was implemented using PyTorch~\cite{paszke2019pytorch} and was trained on the MegaDepth dataset~\cite{li2018megadepth}.
During the training phase, we input images with a size of $832 \times 832$. Our CNN extractor is a deepened ResNet-18~\cite{he2016deep} with features at $1/32$ resolution.
We set the band width $s_0$ in Epipolar Attention to $10$, and the number of anchor point $N_A$ in Structured Feature Extractor is set to $32$.
The matching threshold $\sigma_h$ in Iterable Epipolar Coarse Matching equals to $0.5$, and the threshold in the matching module before refinement is set to $0.2$.
We use $4$ iterartions in  Iterative Epipolar Coarse Matching module.
Our network is trained for 15 epochs with a batch size of $8$, using the Adam~\cite{kingma2014adam} optimizer with an initial learning rate of $1 \times 10^{-3}$.

\begin{table}[ht]
	\centering
	\small
	\caption{Evaluation on HPatches~\cite{balntas2017hpatches} for homography estimation.}
	\label{tab:HPatches_result}
	\vspace{-3mm}
	\scalebox{0.75}{
		\begin{tabular}{c l c c c c}
			\hline
            \multirow{2}{*}{Category} &\multicolumn{1}{c}{\multirow{2}{*}{Method}} &\multicolumn{3}{c}{Homography est. AUC} &\multirow{2}{*}{matches} \\
			\cline{3-5}
                    & &{@3px}  &{@5px}  &{@10px} \\
			\hline
            \multirow{5}{*}{Detector-based} &D2Net~\cite{dusmanu2019d2}+NN &23.2 &35.9 &53.6 &0.2K \\

			                                &R2D2~\cite{r2d2}+NN &50.6 &63.9 &76.8 &0.5K \\

                                            &DISK~\cite{tyszkiewicz2020disk}+NN &52.3 &64.9 &78.9 &1.1K \\

                                            &SP~\cite{detone2018superpoint}+SuperGlue~\cite{sarlin2020superglue} &53.9 &68.3 &81.7 &0.6K \\
                                            &Patch2Pix~\cite{zhou2021patch2pix} &46.4 &59.2 &73.1 &1.0K \\
            \hline
            \multirow{6}{*}{Detector-free} &Sparse-NCNet~\cite{rocco2020efficient} &48.9 &54.2 &67.1 &1.0K \\ 

                                            &COTR~\cite{jiang2021cotr} &41.9 &57.7 &74.0 &1.0K \\
                                            &DRC-Net~\cite{li20dualrc} &50.6 &56.2 &68.3 &1.0K \\
                                            &LoFTR~\cite{sun2021loftr} &65.9 &75.6 &84.6 &1.0K \\
                                            &PDC-Net+~\cite{truong2021pdc} &66.7 &76.8 &85.8 &1.0k \\
			                                &\textbf{SEM(ours)}       &\bf 69.6 &\bf 79.0  &\bf87.1  &1.0K   \\
			\hline
		\end{tabular}
	}
 \vspace{-3mm}
\end{table}

\subsection{Homography Estimation}\label{4.2}
\textbf{Dataset and Metric.}
HPatches~\cite{balntas2017hpatches} is a widely used benchmark for evaluating image matching algorithms. In line with the methodology presented in~\cite{dusmanu2019d2}, we have selected 56 sequences featuring significant viewpoint changes and 52 sequences with substantial illumination variation to assess the performance of our SEM, which was trained on the MegaDepth~\cite{li2018megadepth} dataset. We have adopted the same evaluation protocol used in the LoFTR approach~\cite{sun2021loftr}. In terms of metrics, we report the area under the cumulative curve (AUC) of the corner error distance up to 3, 5, and 10 pixels. In addition, we have limited the maximum number of output matches to 1,000 as LoFTR~\cite{sun2021loftr}.

\textbf{Results.}
Table~\ref{tab:HPatches_result} shows that our SEM sets a new state-of-the-art and notably performs better on HPatches~\cite{balntas2017hpatches} under all error threshold. This is a strong demenstration of the effectiveness of our method.
SEM outperforms the previous state-of-the-art method by a margin of $\textbf{2.9\%}$, $\textbf{2.2\%}$ and $\textbf{1.3\%}$ under 3, 5, 10 pixels respectively.
This shows that our proposed Structured Feature Extractor can effectively model the geometric prior of the image content under illumination changes, while Epipolar Attention/Matching can effectively filter out the effect of irrelevant regions due to the viewpoint variations.


\subsection{Relative Pose Estimation}\label{4.3}
\textbf{Dataset and Metric.}
To evaluate the performance of our SEM in relative pose estimation, we conduct experiments on two datasets, MegaDepth~\cite{li2018megadepth} and ScanNet~\cite{dai2017scannet}.
MegaDepth~\cite{li2018megadepth} is a large-scale outdoor dataset consisting 1 million internet images of 196 different outdoor scenes, reconstructed by COLMAP~\cite{schonberger2016structure}.
Depth maps as intermediate results can be used to obtain ground truth matches.
We follow the same testing pairs and use the same 1500 image pairs as LoFTR~\cite{sun2021loftr}.
All test images are resized to 1216 in their longer dimension.
On the other hand, ScanNet~\cite{dai2017scannet} is usually used to validate the performance of indoor pose estimation, which is composed of monocular sequences with ground truth poses and depth maps.
Wide baselines and extensive textureless regions in image pairs make ScanNet benchmark challenging. 
All test images are resized to $640 \times 480$.
It is worth noting that we evaluate our SEM trained on MegaDepth\cite{li2018megadepth} on ScanNet~\cite{dai2017scannet}.
We report the AUC of pose error at thresholds $(5^{\circ}, 10^{\circ}, 20^{\circ})$ in both benchmark as LoFTR~\cite{sun2021loftr}.
Our SEM with proposed modules achieves state-of-the-art performance on both datasets.

\begin{table}[t]
	\centering
	\small
	\caption{Evaluation on MegaDepth~\cite{li2018megadepth} for outdoor relative position estimation.}
	\label{tab:MegaDepth_result}
	\vspace{-3mm}
	\scalebox{0.75}{
		\begin{tabular}{c l c c c c}
			\hline
            \multirow{2}{*}{Category} &\multicolumn{1}{c}{\multirow{2}{*}{Method}} &\multicolumn{3}{c}{Pose estimation AUC} \\
			\cline{3-5}
                    & &{@$5^\circ$}  &{@$10^\circ$}  &{@$20^\circ$} \\
			\hline
            \multirow{2}{*}{Detector-based} &SP~\cite{detone2018superpoint}+SuperGlue~\cite{sarlin2020superglue} &42.2 &59.0 &73.6 \\
            
                                            &SP~\cite{detone2018superpoint}+SGMNet~\cite{chen2021learning} &40.5 &59.0 &73.6 \\
            \hline
            \multirow{7}{*}{Detector-free}  &DRC-Net~\cite{li20dualrc} &27.0 &42.9 &58.3 \\
                                            
                                            &PDC-Net+(H)~\cite{truong2021pdc}  &43.1  &61.9  &76.1 \\
                                            
                                            &LoFTR~\cite{sun2021loftr} &52.8 &69.2 &81.2 \\                          
                                            
                                            &MatchFormer~\cite{wang2022matchformer} &53.3 &69.7 &81.8 \\
                                            
                                            &QuadTree~\cite{tang2022quadtree} & 54.6 & 70.5 & 82.2 \\
                                            
                                            &ASpanFormer~\cite{chen2022aspanformer} &55.3 &71.5 &83.1 \\

			                                &\textbf{SEM(ours)}       &\textbf{58.0} & \textbf{72.9} &\ \textbf{83.7}  \\
			\hline
		\end{tabular}
	}
\vspace{-3mm}
\end{table}

\begin{table}[t]
	\centering
	\small
	\caption{Evaluation on ScanNet~\cite{dai2017scannet} for indoor relative position estimation. * indicates models trained on MegaDepth~\cite{li2018megadepth}.}
	\label{tab:ScanNet_result}
	\vspace{-3mm}
	\scalebox{0.75}{
		\begin{tabular}{c l c c c c}
			\hline
            \multirow{2}{*}{Category} &\multicolumn{1}{c}{\multirow{2}{*}{Method}} &\multicolumn{3}{c}{Pose estimation AUC} \\
			\cline{3-5}
                    & &{@$5^\circ$}  &{@$10^\circ$}  &{@$20^\circ$} \\
			\hline
            \multirow{3}{*}{Detector-based} &D2-Net~\cite{dusmanu2019d2}+NN &5.3 &14.5 &28.0 \\

											&SP~\cite{detone2018superpoint}+OANet~\cite{zhang2019learning} &11.8 &26.9 &43.9 \\
											
											&SP~\cite{detone2018superpoint}+SuperGlue~\cite{sarlin2020superglue} &16.2 &33.8 &51.8 \\
            \hline
            \multirow{5}{*}{Detector-free}  &DRC-Net~\cite{li20dualrc}* &7.7 &17.9 &30.5 \\                         
            
                                            &MatchFormer~\cite{wang2022matchformer}* &15.8  &32.0   &48.0 \\

											&LoFTR-OT~\cite{sun2021loftr}* &16.9 &33.6 &50.6 \\
                                            

			                                &\textbf{SEM(ours)*}       & \textbf{18.7} & \textbf{36.6} & \textbf{52.9}  \\
			\hline
		\end{tabular}
	}
\end{table}

\begin{figure}[t]
    \centering
    \includegraphics[width=1.0\linewidth]{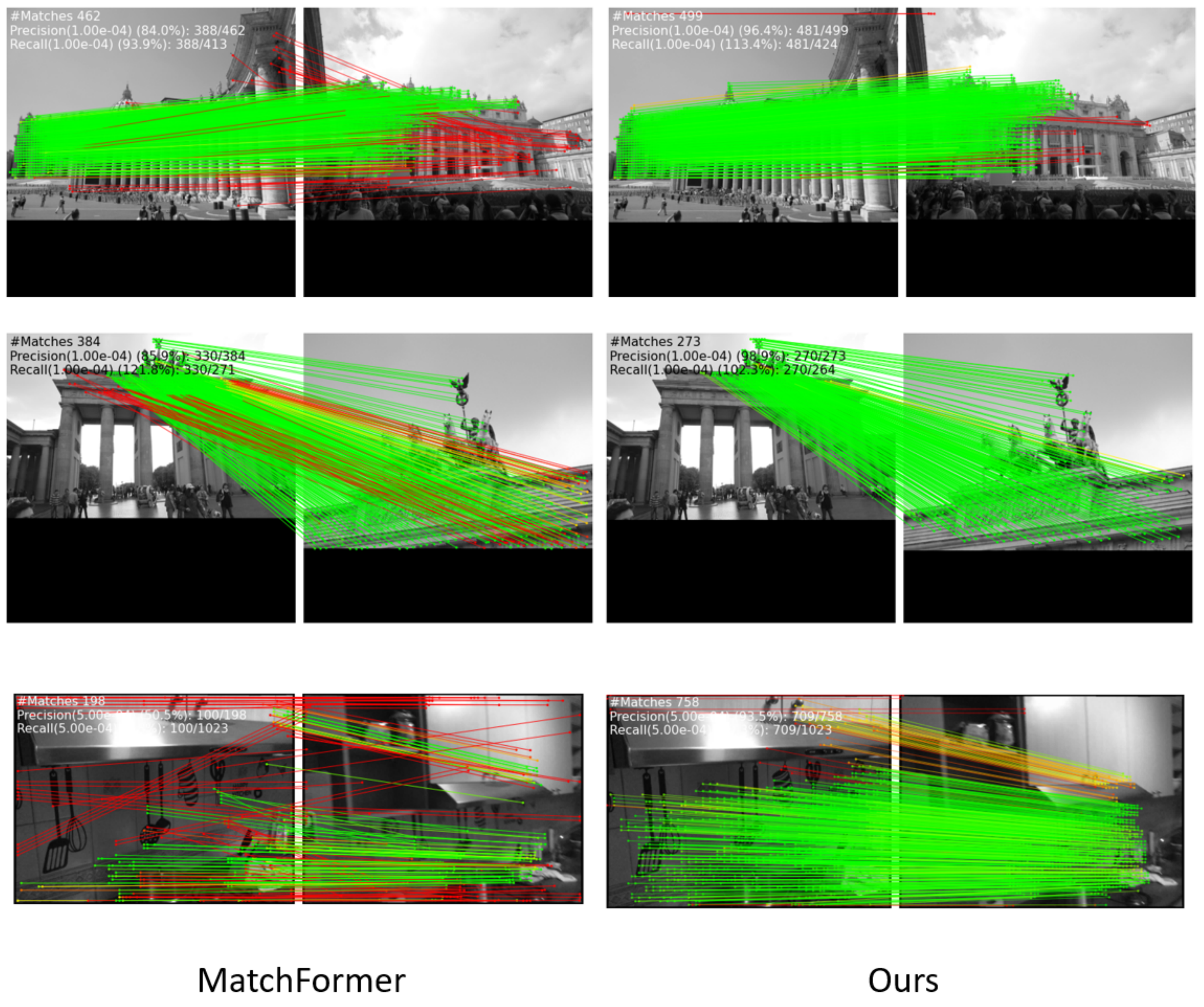}
    \vspace{-6mm}
	\caption{
        Visual qualitative comparison of MatchFormer~\cite{wang2022matchformer} and our SEM on MegaDepth~\cite{li2018megadepth} dataset.
    }\label{fig:qualitative}
    \vspace{-3mm}
\end{figure}

\textbf{Results.}
As Table~\ref{tab:MegaDepth_result} illustrated, our SEM show an outstanding performance on MegaDepth~\cite{li2018megadepth} compared to other methods.
Respectively, our method outperformances the previous best method with $\textbf{2.7\%}$ and $\textbf{1.4\%}$ in AUC$@5^{\circ}$ and AUC$@10^{\circ}$.
This result shows the outstanding performance of our method on outdoor scenes.
Furthermore, Table~\ref{tab:ScanNet_result} compares the performance of our proposed SEM with other state-of-the-art models on ScanNet~\cite{dai2017scannet} dataset.
Similar with MegaDepth~\cite{li2018megadepth}, our method ranks the first although the model is not trained on ScanNet~\cite{dai2017scannet} dataset, which demonstrates the strong generalization ability of our proposed SEM.
The main challenge of indoor dataset lies in the widespread presence of textureless and repetitive texture regions, where structured information is critical.
Our method generalizes well in indoor dataset thanks to the utilization of geometry prior in proposed Structured Feature Extractor and Epipolar Attention and Matching.
In order to more strongly demonstrate the effectiveness of our proposed SEM, Figure~\ref{fig:qualitative} provides a qualitative result compared with other methods. Our method has an outstanding performance on textureless and reppetitive texture regions.


\subsection{Visual Localization}\label{4.4}
\textbf{Dataset and Metric.}
To evaluate the performance of our SEM in visual localization, we use the InLoc~\cite{taira2018inloc} dataset, which consists of 9972 RGBD images.
Among them, 329 RGB images are selected as queries for visual localization.
InLoc~\cite{taira2018inloc} presents challenges such as textureless regions and repetitive patterns under large viewpoint changes.
We evaluate the performance of our SEM trained on MegaDepth~\cite{li2018megadepth} in the same way as LoFTR~\cite{sun2021loftr}.
The metric used in InLoc~\cite{taira2018inloc} measures the percentage of images registered within given error thresholds.

\begin{table}[t]
	\centering
	\small
	\caption{Visual localization evaluation on the InLoc~\cite{taira2018inloc} benchmark.}
	\label{tab:Inloc_result}
	\vspace{-3mm}
	\scalebox{0.75}{
		\begin{tabular}{l c c}
			\hline
            \multicolumn{1}{c}{\multirow{2}{*}{Method}} & DUC1 & DUC2 \\
			\cline{2-3}
                     & \multicolumn{2}{c}{$\left(0.25m, 10^\circ\right)$ / $\left(0.5m, 10^\circ\right)$ / $\left(1m, 10^\circ\right)$} \\
			\hline
            LoFTR~\cite{sun2021loftr} & 47.5 / 72.2 / 84.8 & 54.2 / 74.8 / \textbf{85.5} \\
            MatchFormer~\cite{wang2022matchformer} & 46.5 / 73.2 / 85.9 & \textbf{55.7} / 71.8 / 81.7 \\
            ASpanFormer~\cite{chen2022aspanformer} & 51.5 / 73.7 / 86.4 & 55.0 / 74.0 / 81.7 \\
            \textbf{SEM(ours)} & \textbf{52.0} / \textbf{74.2} / \textbf{87.4} & 50.4 / \textbf{76.3} / 83.2 \\
			\hline
		\end{tabular}
	}
\end{table}

\textbf{Results.}
Our method's performance on the InLoc~\cite{taira2018inloc} benchmark is summarized in Table~\ref{tab:Inloc_result}, where it achieves the best performance on DUC1 and performs comparably to state-of-the-art methods on DUC2.
This demonstrates our method's strong generalization ability in visual localization, even in challenging indoor scenes.

\subsection{Ablation Study}\label{4.5}

\begin{table}[t]
	\centering
	\small
	\caption{Ablation Study of each component on MegaDepth~\cite{li2018megadepth}. SF indicates Structured Feature extractor, and EAM refers to Epipolar Attention and Matching.}
	\label{tab:ablation_study}
	\vspace{-3mm}
	\scalebox{0.75}{
		\begin{tabular}{c c c c c c c}
    		\hline
			\multirow{2}{*}{Index} & \multirow{2}{*}{Multi-Level} & \multirow{2}{*}{SF} & \multirow{2}{*}{EAM}
		    &\multicolumn{3}{c}{Pose estimation AUC} \\
			\cline{5-7}
                & & & & {@$5^\circ$}  &{@$10^\circ$}  &{@$20^\circ$} \\
		    \hline
		      1 & & & & 45.6 & 62.2 & 75.3 \\
		      2 & \checkmark & & & 46.7 & 63.1 & 76.3  \\
		      3 & \checkmark & \checkmark & & 47.3 & 64.3 & 76.8 \\
		      4 & \checkmark & \checkmark & \checkmark & \bf 48.1  & \bf 64.7 & \bf 77.4 \\
		     \hline
		\end{tabular}
	}
\end{table}

To deeply analyze and evaluate the effectiveness of each component in SEM, we conducted detailed ablation studies on MegaDepth~\cite{li2018megadepth}.
Here, we use images with a size of 544 for both training and evaluation.
In Table~\ref{tab:ablation_study}, we gradually added the proposed components to the baseline to analyze their impact.

\begin{figure}[t]
    \centering
    \includegraphics[width=1.0\linewidth]{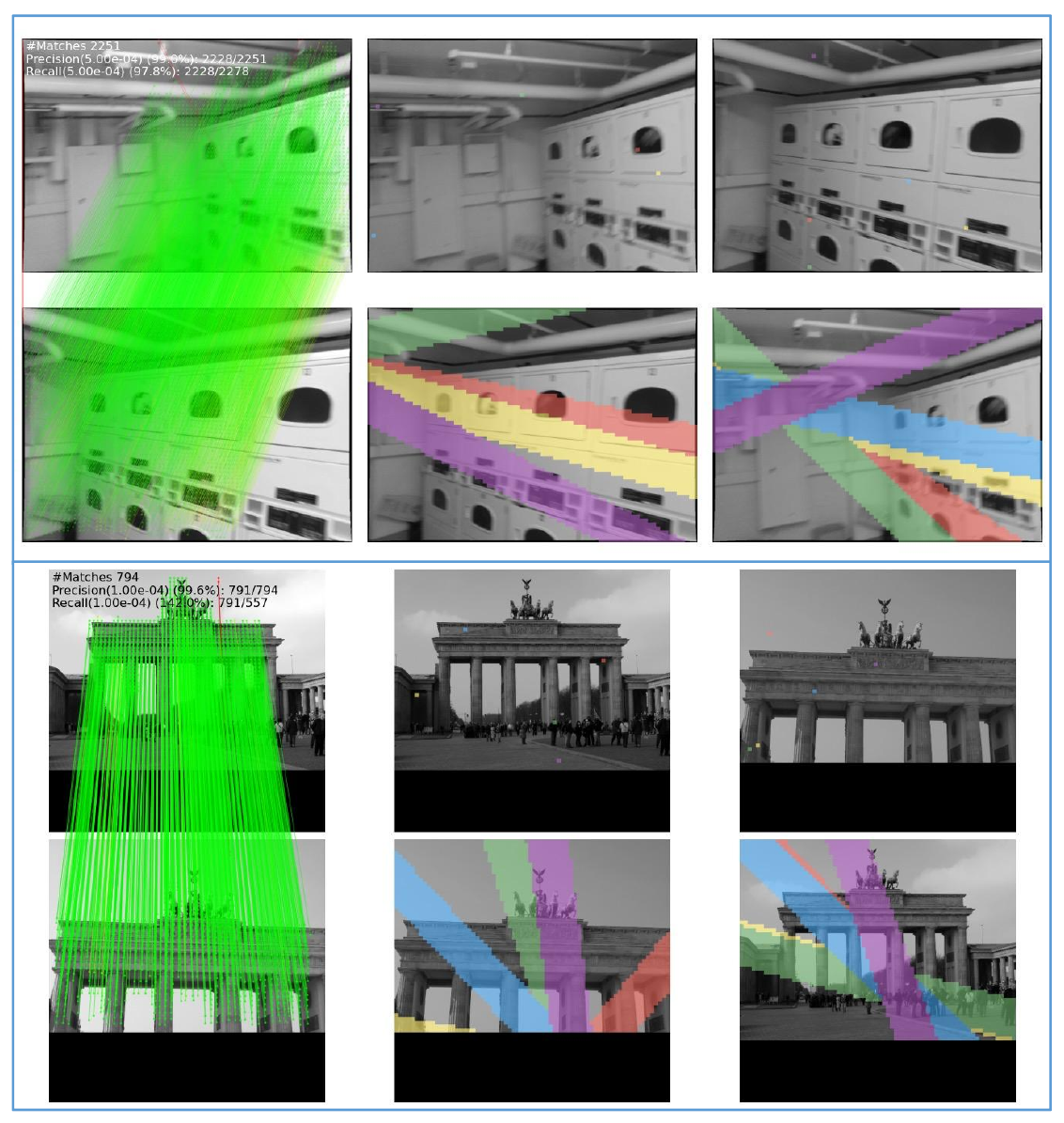}
    \vspace{-6mm}
	\caption{
        Visualization of the epipolar banded areas corresponding to some points and matching results on MegaDepth~\cite{li2018megadepth} and ScanNet~\cite{dai2017scannet}.
        Points and corresponding epipolar regions are identified by the same color.
    }\label{fig:vis_exp}
\vspace{-3mm}
\end{figure}

\textbf{Effectiveness of Structured Feature Extractor.}
According to ~\cite{chen2022aspanformer}, we learn that the multi-level strategy is useful in local matching, which is verfied by the results of Index-2 and Index-1.
Comparing the results of Index-2 and Index-1, we find that the proposed Structured Feature Extractor helps during the inference phase, increasing performance with an improvement of
+0.6 AUC@$5^\circ$, +1.2 AUC@$10^\circ$ and +0.5 AUC@$20^\circ$.

\begin{table}[t]
	\centering
	\small
	\caption{Ablation Study with different $s_0$ in Epipolar Attention on MegaDepth~\cite{li2018megadepth}.}
	\label{tab:s_sigma_result}
	\vspace{-3mm}
	\scalebox{0.85}{
		\begin{tabular}{c c c c}
    		\hline
			\multirow{2}{*}{$s_0$} &\multicolumn{3}{c}{Pose estimation AUC} \\
			\cline{2-4}
                & {@$5^\circ$}  &{@$10^\circ$}  &{@$20^\circ$} \\
            \hline
            5 & 45.6 &  62.7 & 76.2 \\
            10 & \textbf{48.1} & \textbf{64.7} & \textbf{77.4} \\
            15 & 47.5 & 64.3 & 77.2 \\
            20 & 46.7 & 62.4 & 76.4 \\
		     \hline
		\end{tabular}
	}
        \vspace{-3mm}
\end{table}


\textbf{Effectiveness of Epipolar Attention.}
As shown in Table~\ref{tab:ablation_study}, We replace the original linear attention and global matching with our Epipolar Attention and Epipolar Matching in Index-4.
Comparing the results of Index-4 and Index-3, we can see that the performance is improved, which indicates the effectiveness of our proposed Epipolar Attention and Epipolar Matching.
In Figure~\ref{fig:vis_exp}, we visualize the matching results for several pair of images. Specifically, we randomly select points in the reference image and identify their corresponding epipolar band regions (marked by same color) in the source image.
For the points in the source image, the corresponding area is also found in the reference image.
The attention is computed within these points and their corresponding band regions.
The visualization results show that our module can effectively utilize the relative camera poses to filter out irrelevant regions.
To further explore the effect of eliminating irrelevant regions, we explored the effect of different $s_0$ on MegaDepth dataset.
As Table~\ref{tab:s_sigma_result} shows, an appropriate value for $s_0$ is crucial as it can balance the filtering of unwanted regions and tolerance to camera pose estimation errors.
If $s_0$ is too small, it may lead to the exclusion of correct matches when the initial camera pose estimation is imprecise.
On the other hand, if $s_0$ is too large, it may not effectively filter out irrelevant regions.
This shows that the modules we designed play the role we expected.

%% file: conclusion.tex
In this paper, we propose a novel Structured Epipolar Matcher (SEM) for local feature matching utilizing the geometry priors.
To make features more discriminating, we design a Structured Feature Extractor to complement the appearance features.
To exclude the regions of interference as much as possible, we propose Epipolar Attention and Matching in 
 the iterable coarse matching stage.
Extensive experimental results on five benchmarks demonstrate the effectiveness of the proposed method.

%% file: CVPR2023_ASTR.bbl
\begin{thebibliography}{10}\itemsep=-1pt

\bibitem{balntas2017hpatches}
Vassileios Balntas, Karel Lenc, Andrea Vedaldi, and Krystian Mikolajczyk.
\newblock Hpatches: A benchmark and evaluation of handcrafted and learned local
  descriptors.
\newblock In {\em Proceedings of the IEEE Conference on Computer Vision and
  Pattern Recognition}, pages 5173--5182, 2017.

\bibitem{barroso2019key}
Axel Barroso-Laguna, Edgar Riba, Daniel Ponsa, and Krystian Mikolajczyk.
\newblock Key. net: Keypoint detection by handcrafted and learned cnn filters.
\newblock In {\em Proceedings of the IEEE International Conference on Computer
  Vision}, pages 5836--5844, 2019.

\bibitem{berton2021viewpoint}
Gabriele Berton, Carlo Masone, Valerio Paolicelli, and Barbara Caputo.
\newblock Viewpoint invariant dense matching for visual geolocalization.
\newblock In {\em Proceedings of the IEEE/CVF International Conference on
  Computer Vision}, pages 12169--12178, 2021.

\bibitem{chen2021learning}
Hongkai Chen, Zixin Luo, Jiahui Zhang, Lei Zhou, Xuyang Bai, Zeyu Hu, Chiew-Lan
  Tai, and Long Quan.
\newblock Learning to match features with seeded graph matching network.
\newblock In {\em Proceedings of the IEEE/CVF International Conference on
  Computer Vision}, pages 6301--6310, 2021.

\bibitem{chen2022aspanformer}
Hongkai Chen, Zixin Luo, Lei Zhou, Yurun Tian, Mingmin Zhen, Tian Fang, David
  McKinnon, Yanghai Tsin, and Long Quan.
\newblock Aspanformer: Detector-free image matching with adaptive span
  transformer.
\newblock In {\em Computer Vision--ECCV 2022: 17th European Conference, Tel
  Aviv, Israel, October 23--27, 2022, Proceedings, Part XXXII}, pages 20--36.
  Springer, 2022.

\bibitem{dai2017scannet}
Angela Dai, Angel~X Chang, Manolis Savva, Maciej Halber, Thomas Funkhouser, and
  Matthias Nie{\ss}ner.
\newblock Scannet: Richly-annotated 3d reconstructions of indoor scenes.
\newblock In {\em Proceedings of the IEEE conference on computer vision and
  pattern recognition}, pages 5828--5839, 2017.

\bibitem{dai2017bundlefusion}
Angela Dai, Matthias Nie{\ss}ner, Michael Zollh{\"o}fer, Shahram Izadi, and
  Christian Theobalt.
\newblock Bundlefusion: Real-time globally consistent 3d reconstruction using
  on-the-fly surface reintegration.
\newblock {\em ACM Transactions on Graphics (ToG)}, 36(4):1, 2017.

\bibitem{detone2018superpoint}
Daniel DeTone, Tomasz Malisiewicz, and Andrew Rabinovich.
\newblock Superpoint: Self-supervised interest point detection and description.
\newblock In {\em Proceedings of the IEEE Conference on Computer Vision and
  Pattern Recognition Workshops}, pages 224--236, 2018.

\bibitem{dusmanu2019d2}
Mihai Dusmanu, Ignacio Rocco, Tomas Pajdla, Marc Pollefeys, Josef Sivic,
  Akihiko Torii, and Torsten Sattler.
\newblock D2-net: A trainable cnn for joint detection and description of local
  features.
\newblock In {\em Proceedings of the IEEE Conference on Computer Vision and
  Pattern Recognition}, 2019.

\bibitem{edstedt2022deep}
Johan Edstedt, M{\aa}rten Wadenb{\"a}ck, and Michael Felsberg.
\newblock Deep kernelized dense geometric matching.
\newblock {\em arXiv preprint arXiv:2202.00667}, 2022.

\bibitem{germain2021visual}
Hugo Germain, Vincent Lepetit, and Guillaume Bourmaud.
\newblock Visual correspondence hallucination.
\newblock {\em arXiv preprint arXiv:2106.09711}, 2021.

\bibitem{grabner20183d}
Alexander Grabner, Peter~M Roth, and Vincent Lepetit.
\newblock 3d pose estimation and 3d model retrieval for objects in the wild.
\newblock In {\em Proceedings of the IEEE Conference on Computer Vision and
  Pattern Recognition}, pages 3022--3031, 2018.

\bibitem{he2016deep}
Kaiming He, Xiangyu Zhang, Shaoqing Ren, and Jian Sun.
\newblock Deep residual learning for image recognition.
\newblock In {\em Proceedings of the IEEE Conference on Computer Vision and
  Pattern Recognition}, pages 770--778, 2016.

\bibitem{horn1981determining}
Berthold~KP Horn and Brian~G Schunck.
\newblock Determining optical flow.
\newblock {\em Artificial intelligence}, 17(1-3):185--203, 1981.

\bibitem{huang2019dynamic}
Shuaiyi Huang, Qiuyue Wang, Songyang Zhang, Shipeng Yan, and Xuming He.
\newblock Dynamic context correspondence network for semantic alignment.
\newblock In {\em Proceedings of the IEEE International Conference on Computer
  Vision}, pages 2010--2019, 2019.

\bibitem{jiang2021cotr}
Wei Jiang, Eduard Trulls, Jan Hosang, Andrea Tagliasacchi, and Kwang~Moo Yi.
\newblock Cotr: Correspondence transformer for matching across images.
\newblock In {\em Proceedings of the IEEE/CVF International Conference on
  Computer Vision}, pages 6207--6217, 2021.

\bibitem{katharopoulos2020transformers}
Angelos Katharopoulos, Apoorv Vyas, Nikolaos Pappas, and Fran{\c{c}}ois
  Fleuret.
\newblock Transformers are rnns: Fast autoregressive transformers with linear
  attention.
\newblock In {\em International Conference on Machine Learning}, pages
  5156--5165. PMLR, 2020.

\bibitem{kingma2014adam}
Diederik~P Kingma and Jimmy Ba.
\newblock Adam: A method for stochastic optimization.
\newblock {\em arXiv preprint arXiv:1412.6980}, 2014.

\bibitem{li20dualrc}
Xinghui Li, Kai Han, Shuda Li, and Victor Prisacariu.
\newblock Dual-resolution correspondence networks.
\newblock {\em Advances in Neural Information Processing Systems}, 33, 2020.

\bibitem{li2018megadepth}
Zhengqi Li and Noah Snavely.
\newblock Megadepth: Learning single-view depth prediction from internet
  photos.
\newblock In {\em Proceedings of the IEEE Conference on Computer Vision and
  Pattern Recognition}, pages 2041--2050, 2018.

\bibitem{lin2017feature}
Tsung-Yi Lin, Piotr Doll{\'a}r, Ross Girshick, Kaiming He, Bharath Hariharan,
  and Serge Belongie.
\newblock Feature pyramid networks for object detection.
\newblock In {\em Proceedings of the IEEE Conference on Computer Vision and
  Pattern Recognition}, pages 2117--2125, 2017.

\bibitem{lowe2004distinctive}
David~G Lowe.
\newblock Distinctive image features from scale-invariant keypoints.
\newblock {\em International Journal of Computer Vision}, 60(2):91--110, 2004.

\bibitem{lucas1981iterative}
Bruce~D Lucas, Takeo Kanade, et~al.
\newblock {\em An iterative image registration technique with an application to
  stereo vision}, volume~81.
\newblock Vancouver, 1981.

\bibitem{ono2018lf}
Yuki Ono, Eduard Trulls, Pascal Fua, and Kwang~Moo Yi.
\newblock Lf-net: learning local features from images.
\newblock In {\em Advances in Neural Information Processing Systems}, pages
  6237--6247, 2018.

\bibitem{parihar2021rord}
Udit~Singh Parihar, Aniket Gujarathi, Kinal Mehta, Satyajit Tourani, Sourav
  Garg, Michael Milford, and K~Madhava Krishna.
\newblock Rord: Rotation-robust descriptors and orthographic views for local
  feature matching.
\newblock In {\em 2021 IEEE/RSJ International Conference on Intelligent Robots
  and Systems (IROS)}, pages 1593--1600. IEEE, 2021.

\bibitem{paszke2019pytorch}
Adam Paszke, Sam Gross, Francisco Massa, Adam Lerer, James Bradbury, Gregory
  Chanan, Trevor Killeen, Zeming Lin, Natalia Gimelshein, Luca Antiga, et~al.
\newblock Pytorch: An imperative style, high-performance deep learning library.
\newblock {\em Advances in neural information processing systems}, 32, 2019.

\bibitem{persson2018lambda}
Mikael Persson and Klas Nordberg.
\newblock Lambda twist: An accurate fast robust perspective three point (p3p)
  solver.
\newblock In {\em Proceedings of the European Conference on Computer Vision},
  pages 318--332, 2018.

\bibitem{r2d2}
Jerome Revaud, Philippe Weinzaepfel, C{\'{e}}sar~Roberto de Souza, and Martin
  Humenberger.
\newblock {R2D2:} repeatable and reliable detector and descriptor.
\newblock In {\em Advances in Neural Information Processing Systems}, 2019.

\bibitem{rocco2020efficient}
Ignacio Rocco, Relja Arandjelovi{\'c}, and Josef Sivic.
\newblock Efficient neighbourhood consensus networks via submanifold sparse
  convolutions.
\newblock In {\em Proceedings of the European Conference on Computer Vision},
  pages 605--621, 2020.

\bibitem{rocco2018neighbourhood}
Ignacio Rocco, Mircea Cimpoi, Relja Arandjelovi{\'c}, Akihiko Torii, Tomas
  Pajdla, and Josef Sivic.
\newblock Neighbourhood consensus networks.
\newblock In {\em Advances in Neural Information Processing Systems}, pages
  1658--1669, 2018.

\bibitem{rublee2011orb}
Ethan Rublee, Vincent Rabaud, Kurt Konolige, and Gary Bradski.
\newblock Orb: An efficient alternative to sift or surf.
\newblock In {\em 2011 International conference on computer vision}, pages
  2564--2571. Ieee, 2011.

\bibitem{sarlin2020superglue}
Paul-Edouard Sarlin, Daniel DeTone, Tomasz Malisiewicz, and Andrew Rabinovich.
\newblock Superglue: Learning feature matching with graph neural networks.
\newblock In {\em Proceedings of the IEEE Conference on Computer Vision and
  Pattern Recognition}, pages 4938--4947, 2020.

\bibitem{sattler2018benchmarking}
Torsten Sattler, Will Maddern, Carl Toft, Akihiko Torii, Lars Hammarstrand,
  Erik Stenborg, Daniel Safari, Masatoshi Okutomi, Marc Pollefeys, Josef Sivic,
  et~al.
\newblock Benchmarking 6dof outdoor visual localization in changing conditions.
\newblock In {\em Proceedings of the IEEE Conference on Computer Vision and
  Pattern Recognition}, pages 8601--8610, 2018.

\bibitem{schonberger2016structure}
Johannes~L Schonberger and Jan-Michael Frahm.
\newblock Structure-from-motion revisited.
\newblock In {\em Proceedings of the IEEE Conference on Computer Vision and
  Pattern Recognition}, pages 4104--4113, 2016.

\bibitem{sun2021loftr}
Jiaming Sun, Zehong Shen, Yuang Wang, Hujun Bao, and Xiaowei Zhou.
\newblock Loftr: Detector-free local feature matching with transformers.
\newblock In {\em Proceedings of the IEEE/CVF Conference on Computer Vision and
  Pattern Recognition}, pages 8922--8931, 2021.

\bibitem{sun2020acne}
Weiwei Sun, Wei Jiang, Eduard Trulls, Andrea Tagliasacchi, and Kwang~Moo Yi.
\newblock Acne: Attentive context normalization for robust
  permutation-equivariant learning.
\newblock In {\em Proceedings of the IEEE/CVF conference on computer vision and
  pattern recognition}, pages 11286--11295, 2020.

\bibitem{sun2023neuralbf}
Weiwei Sun, Daniel Rebain, Renjie Liao, Vladimir Tankovich, Soroosh Yazdani,
  Kwang~Moo Yi, and Andrea Tagliasacchi.
\newblock Neuralbf: Neural bilateral filtering for top-down instance
  segmentation on point clouds.
\newblock In {\em Proceedings of the IEEE/CVF Winter Conference on Applications
  of Computer Vision}, pages 551--560, 2023.

\bibitem{taira2018inloc}
Hajime Taira, Masatoshi Okutomi, Torsten Sattler, Mircea Cimpoi, Marc
  Pollefeys, Josef Sivic, Tomas Pajdla, and Akihiko Torii.
\newblock Inloc: Indoor visual localization with dense matching and view
  synthesis.
\newblock In {\em Proceedings of the IEEE Conference on Computer Vision and
  Pattern Recognition}, pages 7199--7209, 2018.

\bibitem{tang2022quadtree}
Shitao Tang, Jiahui Zhang, Siyu Zhu, and Ping Tan.
\newblock Quadtree attention for vision transformers.
\newblock {\em arXiv preprint arXiv:2201.02767}, 2022.

\bibitem{teed2020raft}
Zachary Teed and Jia Deng.
\newblock Raft: Recurrent all-pairs field transforms for optical flow.
\newblock In {\em Computer Vision--ECCV 2020: 16th European Conference,
  Glasgow, UK, August 23--28, 2020, Proceedings, Part II 16}, pages 402--419.
  Springer, 2020.

\bibitem{toft2020single}
Carl Toft, Daniyar Turmukhambetov, Torsten Sattler, Fredrik Kahl, and Gabriel~J
  Brostow.
\newblock Single-image depth prediction makes feature matching easier.
\newblock In {\em Computer Vision--ECCV 2020: 16th European Conference,
  Glasgow, UK, August 23--28, 2020, Proceedings, Part XVI 16}, pages 473--492.
  Springer, 2020.

\bibitem{truong2020gocor}
Prune Truong, Martin Danelljan, Luc~V Gool, and Radu Timofte.
\newblock Gocor: Bringing globally optimized correspondence volumes into your
  neural network.
\newblock {\em Advances in Neural Information Processing Systems},
  33:14278--14290, 2020.

\bibitem{truong2020glu}
Prune Truong, Martin Danelljan, and Radu Timofte.
\newblock Glu-net: Global-local universal network for dense flow and
  correspondences.
\newblock In {\em Proceedings of the IEEE Conference on Computer Vision and
  Pattern Recognition}, pages 6258--6268, 2020.

\bibitem{truong2021pdc}
Prune Truong, Martin Danelljan, Radu Timofte, and Luc Van~Gool.
\newblock Pdc-net+: Enhanced probabilistic dense correspondence network.
\newblock {\em arXiv preprint arXiv:2109.13912}, 2021.

\bibitem{tyszkiewicz2020disk}
Micha{\l} Tyszkiewicz, Pascal Fua, and Eduard Trulls.
\newblock Disk: Learning local features with policy gradient.
\newblock {\em Advances in Neural Information Processing Systems},
  33:14254--14265, 2020.

\bibitem{wang2022matchformer}
Qing Wang, Jiaming Zhang, Kailun Yang, Kunyu Peng, and Rainer Stiefelhagen.
\newblock Matchformer: Interleaving attention in transformers for feature
  matching.
\newblock In {\em Proceedings of the Asian Conference on Computer Vision},
  pages 2746--2762, 2022.

\bibitem{yi2016lift}
Kwang~Moo Yi, Eduard Trulls, Vincent Lepetit, and Pascal Fua.
\newblock Lift: Learned invariant feature transform.
\newblock In {\em Proceedings of the European Conference on Computer Vision},
  pages 467--483, 2016.

\bibitem{yu2023adaptive}
Jiahuan Yu, Jiahao Chang, Jianfeng He, Tianzhu Zhang, and Feng Wu.
\newblock Adaptive spot-guided transformer for consistent local feature
  matching.
\newblock {\em arXiv preprint arXiv:2303.16624}, 2023.

\bibitem{zhang2019learning}
Jiahui Zhang, Dawei Sun, Zixin Luo, Anbang Yao, Lei Zhou, Tianwei Shen, Yurong
  Chen, Long Quan, and Hongen Liao.
\newblock Learning two-view correspondences and geometry using order-aware
  network.
\newblock In {\em Proceedings of the IEEE International Conference on Computer
  Vision}, pages 5845--5854, 2019.

\bibitem{zhou2021patch2pix}
Qunjie Zhou, Torsten Sattler, and Laura Leal-Taixe.
\newblock Patch2pix: Epipolar-guided pixel-level correspondences.
\newblock In {\em Proceedings of the IEEE/CVF conference on computer vision and
  pattern recognition}, pages 4669--4678, 2021.

\end{thebibliography}
